\documentclass[conference]{ieeeconf}
\IEEEoverridecommandlockouts
\usepackage{graphicx}
\usepackage{cite}
\usepackage{picinpar}
\usepackage{hhline}
\usepackage{amsmath}
\usepackage{url}
\usepackage{booktabs}
\usepackage{caption}
\usepackage{multirow}
\usepackage{graphicx}
\usepackage[latin1]{inputenc}
\usepackage{colortbl}
\usepackage{soul}
\usepackage{multirow}
\usepackage{pifont}
\usepackage{color}
\usepackage{alltt}
\usepackage{enumerate}
\usepackage{siunitx}
\usepackage{epstopdf}
\usepackage{pbox}
\usepackage{amssymb}
\usepackage{mathtools}
\usepackage{bbm}
\usepackage[normalem]{ulem}
\usepackage{cite}
\usepackage{amsfonts}
\usepackage{algorithmic}
\usepackage{textcomp}
\usepackage{xcolor}

\usepackage{caption}

\definecolor{limegreen}{rgb}{0.2, 0.8, 0.2}
\definecolor{forestgreen}{rgb}{0.13, 0.55, 0.13}
\definecolor{greenhtml}{rgb}{0.0, 0.5, 0.0}
\definecolor{black}{rgb}{0.0, 0.0, 0.0}

\usepackage[colorlinks,allcolors=black]{hyperref}

\usepackage[font=small,labelfont=bf,tableposition=top]{caption}

\usepackage{blindtext}
\title{here title}

\let\oldtwocolumn\twocolumn
\renewcommand\twocolumn[1][]{%
    \oldtwocolumn[{#1}{
    \begin{center}
           \includegraphics[width=16cm]{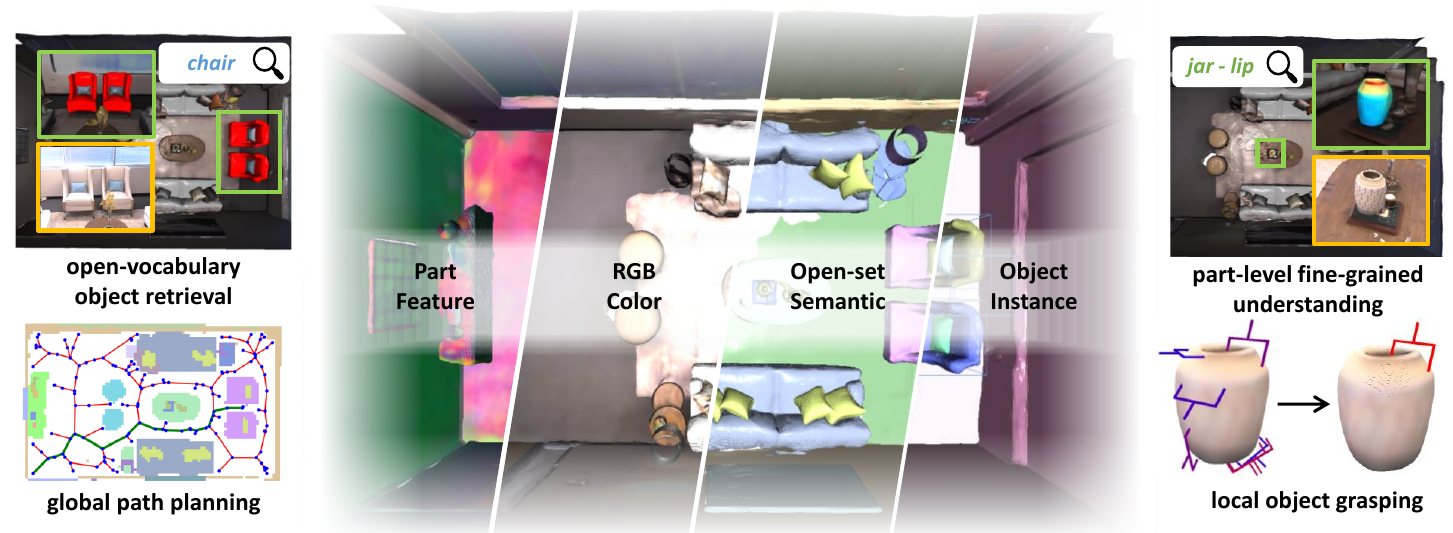}
           \captionof{figure}{We introduce \textbf{OpenObj}, a framework of open-vocabulary object-level neural radiance fields with fine-grained understanding. OpenObj facilitates various downstream tasks, including open-vocabulary object retrieval, part-level fine-grained understanding, zero-shot semantic segmentation, and so on.}
           \label{first}
        \end{center}
    }]
}

\begin{document}

\pagestyle{plain}
\title
{	
\textbf{OpenObj: Open-Vocabulary Object-Level Neural Radiance Fields with Fine-Grained Understanding}
    
}

\author{
Yinan Deng, Jiahui Wang, Jingyu Zhao, Jianyu Dou,  Yi Yang, and Yufeng Yue$^{*}$
\thanks{This work is supported by the National Natural Science Foundation of China under Grant  62003039, 61973034, U193203, 62173042. (Corresponding Author: Yufeng Yue, yueyufeng@bit.edu.cn)}
\thanks{All authors are with School of Automation, Beijing Institute of Technology, Beijing, 100081, China.}
}

\maketitle

\begin{abstract}

In recent years, there has been a surge of interest in open-vocabulary 3D scene reconstruction facilitated by visual language models (VLMs), which showcase remarkable capabilities in open-set retrieval. However, existing methods face some limitations: they either focus on learning point-wise features, resulting in blurry semantic understanding, or solely tackle object-level reconstruction, thereby overlooking the intricate details of the object's interior.
To address these challenges, we introduce OpenObj, an innovative approach to build open-vocabulary object-level Neural Radiance Fields (NeRF) with fine-grained understanding. In essence, OpenObj establishes a robust framework for efficient and watertight scene modeling and comprehension at the object-level. Moreover, we incorporate part-level features into the neural fields, enabling a nuanced representation of object interiors. This approach captures object-level instances while maintaining a fine-grained understanding.
The results on multiple datasets demonstrate that OpenObj achieves superior performance in zero-shot semantic segmentation and retrieval tasks. Additionally, OpenObj supports real-world robotics tasks at multiple scales, including global movement and local manipulation.
The project page of OpenObj is available at \urlstyle{tt} \underline{\url{https://OpenObj.github.io/}}.

\end{abstract}

\section{Introduction}

The accurate reconstruction and comprehensive understanding of a 3D scene are critical for guiding robots in performing downstream tasks. 
Classical map-building strategies, such as Octomap \cite{OctoMap}, focus on reconstructing the geometric structure of the scenes, primarily facilitating obstacle avoidance and fixed-point spatial navigation for robots, e.g., `\textit{Please go to \uline{location (x, y)}}'.  With the advancement of deep learning techniques, some semantic priors are now embedded into maps to support navigation tasks at a semantic level \cite{S-MKI}, e.g., `\textit{Please go to the vicinity of the \uline{table}}'. However, these semantics are limited to a closed-set of labels predefined during the training phase \cite{see-csom}, making it challenging to generalize to new scenes or the real world where concepts and categories are more diverse and abundant.

Recently, Visual Language Models (VLMs) \cite{CLIP, ALIGN, blip} pre-trained on web-scale data have garnered widespread attention for their ability to infer rich semantic knowledge from visual images and their strong generalization capabilities. Leveraging the powerful zero-shot perception of VLMs, numerous open-vocabulary mapping methods have emerged, integrating VLM features with traditional mapping frameworks. These maps facilitate human interaction and support higher-level cognitive navigation,  e.g., `\textit{Please find \uline{a soft piece of furniture}.}' or `\textit{Please find me \uline{a place to rest}}'. 

Most of the current open-vocabulary mapping methods focus on obtaining dense pixel-wise VLM features from 2D visual images and distilling \cite{lerf, 3dovs} or projecting \cite{Clip-fields, Conceptfusion} them into 3D space. However, this approach only yields point-wise features, limiting the utility of these maps due to the absence of object-level understanding. To address this limitation, some works \cite{OV-NeRF, conceptgraphs} have proposed instance-oriented open-vocabulary mapping methods. They often leverage SAM (Segment Anything Model) \cite{sam}, which has a strong ability to extract zero-shot region proposals, as the basis for instance segmentation. However, these methods recognize scenes only at the object level and fail to provide a more granular understanding of internal structures.  The question then arises: \textbf{What is the effective granularity of an open-vocabulary map representation?}

In addressing this challenge, we are inspired by how humans cognitively process their environment. When encountering a new scene, humans first generate a rough representation of the whole (e.g., \textit{`here is a table with several cups on it'}). Upon closer observation, they then derive a detailed description of the individual components of specific objects (e.g., \textit{`this cup has a square handle and a rabbit pattern'}). Following this inspiration, we proposed OpenObj, an innovative approach to build open-vocabulary object-level neural radiance fields with fine-grained understanding. Our key idea is to build an object-level map, where each object is modeled as an independent implicit field to learn photometric, geometric, and part-level features.

Specifically, we perform instance segmentation and object-level understanding on the visual images, and propose a two-stage mask clustering method to ensure segmentation consistency across frames. Then, we leverage the over-segmentation capability of SAM and the image encoding ability of CLIP to obtain 2D dense pixel-level feature embedding, which provides more detailed part-level knowledge. Finally, for each object instance, we construct a NeRF that simultaneously fits the color, geometry, and features. In this way, the resulting map representation offers multi-granularity understanding and watertight reconstruction. At a coarse level, OpenObj enables object-oriented retrieval and navigation; at a fine level, OpenObj supports the representation and manipulation of specific objects.

In summary, Our contributions are summarized as follows:

\begin{itemize}
    \item We present OpenObj, the open-vocabulary object-level neural radiance fields with fine-grained understanding, supporting downstream tasks at multiple scales.
    \item We propose a two-stage mask clustering method to ensure consistent instance segmentation across frames.
    \item We develop a technique for extracting fine-grained part-level VLM features from 2D images.
    \item  Qualitative and quantitative results in multiple scenes demonstrate that OpenObj excels in zero-shot segmentation and open-vocabulary retrieval.
\end{itemize}


\begin{figure*}[!t]\centering
	\includegraphics[width=16.5cm]{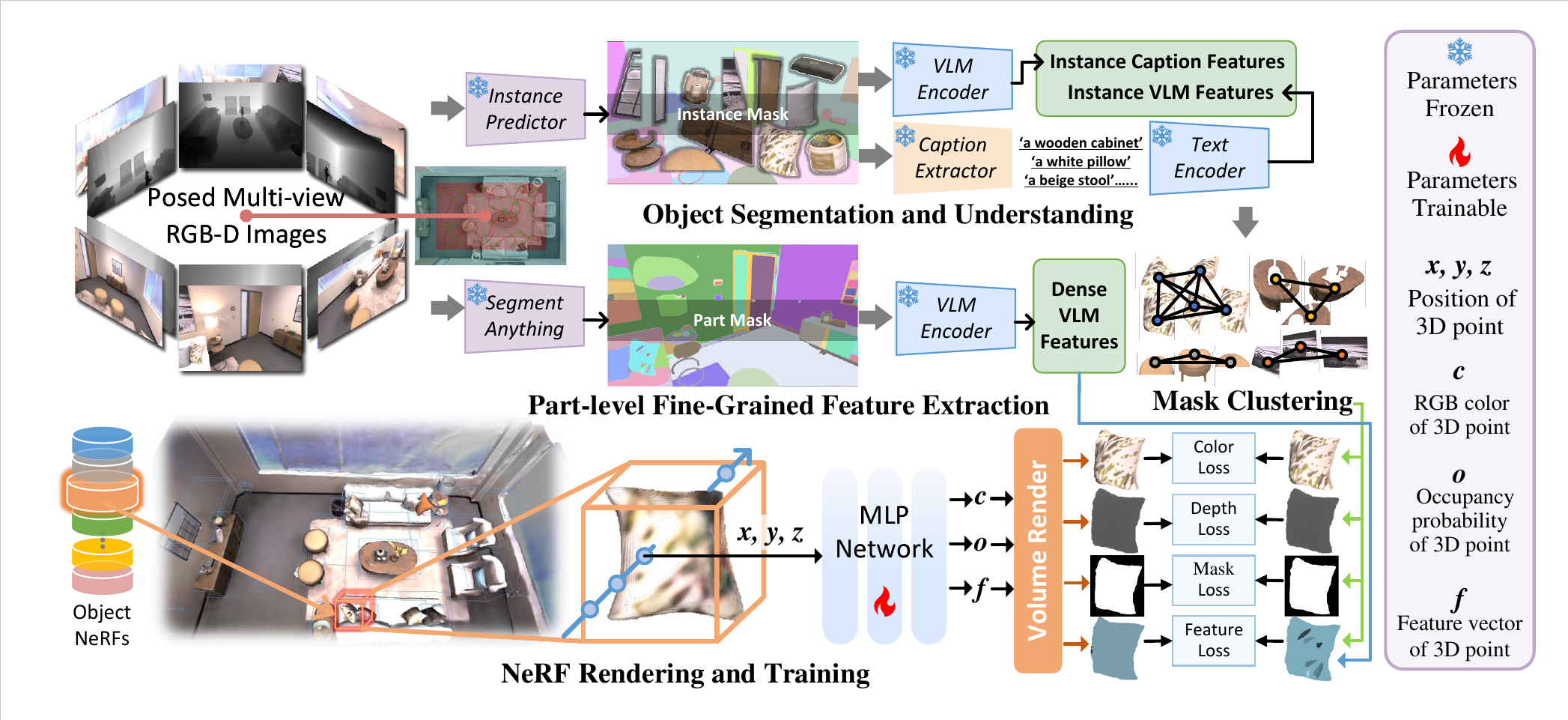}
	\caption{The framework of OpenObj consists of four main modules: Object Segmentation and Understanding, Mask Clustering, Part-level Fine-Grained Feature Extraction, and Hierarchical Graph Representation Formation. } 
	\label{framework} 
\end{figure*}

\section{Related Works} \label{RW}

\subsection{Closed-set semantic mapping}
Advancements in deep learning for 2D and 3D semantic understanding have enabled the integration of semantic data into operable scene models for robots \cite{see-csom, hd-ccsom}. Utilizing close-set semantic cues embedded in representations, especially object-oriented, a wide range of robotic downstream tasks can be effectively carried out \cite{sem-survey}. For example, methods like \cite{objwang} leverage visual and semantic priors captured from stereo cameras and Web Ontology Language (OWL) to share semantic knowledge with robots in constrained indoor environments. Moreover, \cite{meaningful} enables object-oriented semantic mapping leveraging feature-based RGBD SLAM, deep-learning object detection and 3D unsupervised segmentation. Despite the progress in integrating semantics into robotics, the above methods either rely on segmentation models pre-trained with limited class sets or are confined to coarse semantic comprehension. These constraints pose challenges to their practical deployment in real-world contexts.
 
\subsection{Open-vocabulary 3D mapping}
To leverage the zero-shot generalization and visual-language reasoning capabilities of VLMs and LLMs (Large Language Models) for scene understanding and robotic tasks, numerous approaches have been developed.
Early endeavors like ConceptFusion \cite{Conceptfusion} projected RGB-D image features onto 3D point clouds to yield a multi-modal queryable map representation. Similarly, OpenScene \cite{openscene} employs an additional 3D distillation network for direct prediction of visual language features in 3D spaces. However, these approaches lack clear object segmentation, which limits their practical usage for robotic interactive tasks. 

Consistent open-vocabulary instance segmentation across views is vital for object-level scene understanding. ConceptGraphs \cite{conceptgraphs} and OpenGraph \cite{opengraph} address this by iteratively fusing per-frame feature point clouds, leveraging geometric and feature similarity metrics. OpenMask3D  \cite{openmask3d} utilizes predicted class-agnostic 3D instance masks to guide the multi-view fusion of CLIP embeddings.
However, they only comprehend the scene at a general object level, lacking the ability to provide fine-grained part-level understanding in robotic tasks, especially for manipulation.

Recently, the impressive performance of 3D Gaussian Splatting (3DGS) \cite{3DGS} for real-time high-fidelity rendering has been notable in scene representation. Several approaches \cite{langsplat, opengaussian} have tried to integrate 3DGS with VLM features. However, the inherent explicit structure of 3D Gaussian lacks storage efficiency, posing a challenge for achieving fine-grained point-wise understanding. Instead, OpenObj mitigates this problem by employing NeRF models with simple structures.

\subsection{Neural Radiance Fields}
As a compact scene representation for novel view synthesis, NeRF \cite{nerf} and its variants essentially encode 3D scenes in the weights of trainable MLPs or immediate features. Naturally, supervised with close-set semantic segmentation images, Semantic-NeRF \cite{place} can represent the semantic logits of any point in space for novel view label image synthesis. Moreover, open-vocabulary NeRF can be realized by leveraging readily available visual-language features from images as supervised pseudo-truths.

LERF \cite{lerf} initially incorporates multi-scale CLIP features into the NeRF model, though the exhaustive scale search significantly impacts training and inference efficiency. Additionally, 3D-OVS \cite{3dovs} optimized a semantic feature field using an additional relevancy-distribution alignment loss to enhance segmentation.  Furthermore,  OV-NeRF \cite{OV-NeRF} proposes a ranking regularization and a cross-view self-enhancement strategy for denoising and ensuring view consistency, respectively. For integration with robotics missions, CLIP-fields \cite{Clip-fields} introduce the first open-vocabulary neural feature fields as robotic semantic memory. GeFF \cite{GeFF} further integrates 2D VLM features into generalizable NeRF model as a unified representation for both navigation and manipulation. Despite their impressive performance in practical scenarios, they still struggle with representing the entire scene coherently without a clear understanding of objects.

To address this challenge, OpenObj takes the instance detection model as a front-end and leverages a vectorized object mapping approach, inspired by vMAP \cite{vmap}. Along with part segmentation and comprehension, OpenObj builds open-vocabulary object-level NeRFs with fine-grained understanding, enhancing mobile robots' interaction in complex environments.

\section{OpenObj}  \label{method}

\subsection{Framework Overview}

OpenObj processes a series of multi-view color images $\mathcal{I}=\{I^{c}_{1},I^{c}_{2},...,I^{c}_{t}\}$ and depth images $\mathcal{I}=\{I^{d}_{1},I^{d}_{2},...,I^{d}_{t}\}$ with poses $\mathcal{P}=\{P_{1},P_{2},...,P_{t}\}$ collected in a scene, and gradually reconstructs an open-vocabulary map of the scene. This map is organized as a stack of objects, with each element comprising the overall understanding, and a NeRF. The backbone of NeRF is a small MLP (Multilayer Perceptron) that takes 3D point coordinates $x, y, z$, and outputs color $c$, occupancy probability  $\sigma$, and features vector $f$, providing detailed color, geometric, and part-level understanding of the object.

The framework of OpenObj, illustrated in Fig. \ref{framework}, comprises four main modules. First, the Object Segmentation and Understanding module identifies and comprehends object instances from color images. Then, the Mask Clustering module ensures consistent object association across frames. Next, the Part-level Fine-Grained Feature Extraction module leverages the dense segmentation capability of SAM to distinguish parts and extracts their visual features using VLMs. Finally, the NeRF Rendering and Training module vectorizes the training of NeRFs for all objects based on the masks, input RGBD images, and dense VLM features, enabling it to learn detailed object properties.

\subsection{Object Segmentation and Understanding}

Vision, as the primary sense for both humans and robots to perceive the world, provides rich color and texture information essential for understanding the environment. We begin by applying an off-the-shelf class-agnostic mask predictor to each color image $I^{c}_t$, generating a set of 2D masks $\{m^{obj}_{t,i} \mid i=1, 2, \ldots, n^{obj}_t\}$. These masks are expected to be instance-level, meaning that pixels belonging to the same object are grouped into the same mask. The advanced instance segmentation tool CropFormer \cite{cropformer} effectively meets this requirement.

To understand these segmented objects, we need to process the images using foundational models. Through visual and text contrastive learning, VLMs possess open-vocabulary cognition, allowing them to capture multiple attributes of objects, such as color and material. In this paper, we use the visual encoder of CLIP \cite{CLIP} to encode images cropped according to the mask $m^{obj}_{t,i}$ as VLM feature $f^{clip}_{t,i}$. 

Additionally, we apply another method to compensate for the limitations of VLM features  $f^{clip}_{t,i}$ in semantic reasoning. Specifically, we use the bounding boxes of the masks $m^{obj}_{t,i}$ as prompts and use the TAP (Tokenize Anything via Prompting) model \cite{tap} to generate a caption $cap_{t,i}$ for each mask,  which is typically a simple phrase. Given the strong advantages of LLMs in natural language processing tasks, we encode these captions using LLMs to obtain their caption features $f^{cap}_{t,i}$. This approach enhances object understanding by enabling common-sense reasoning through caption text encoding. In this paper, we choose SBERT to implement this process.

In this module, we obtain instance masks $m^{obj}_{t,i}$ as well as their CLIP features $f^{clip}_{t,i}$ and caption features $f^{cap}_{t,i}$.

\subsection{Mask Clustering}\label{MC}

\begin{figure}[!t]\centering
	\includegraphics[width=7.5cm]{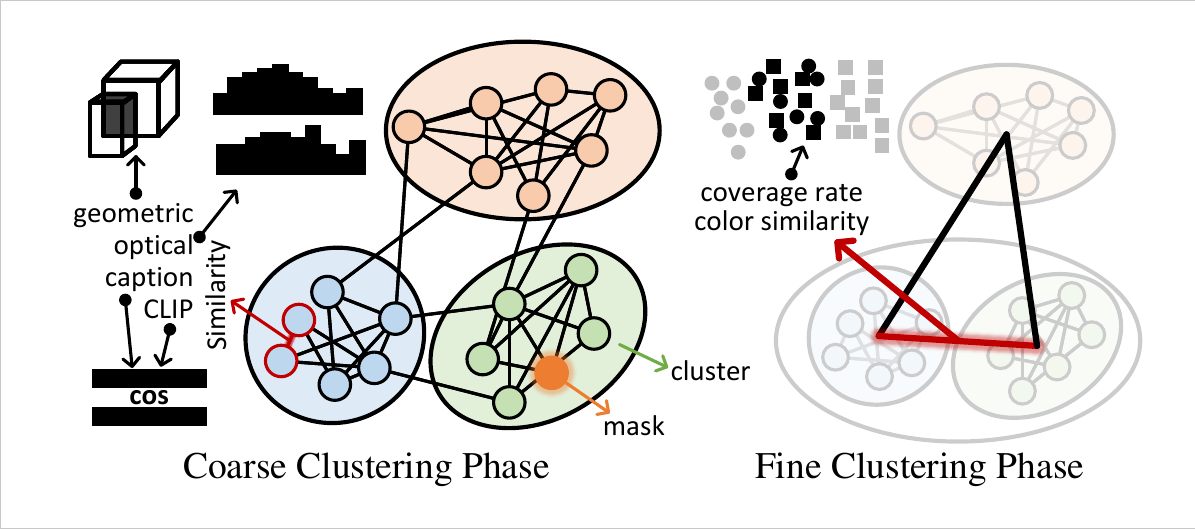}
	\caption{Two-stage mask clustering. In the coarse clustering phase, a graph is constructed for all masks, and the Louvain algorithm is applied to achieve clustering. In the fine clustering stage, the clusters are further fused according to the matched points coverage rate and color similarity of the superimposed point cloud.} 
	\label{clustering} 
\end{figure}

Associating masks belonging to the same object in different frames is crucial for subsequent object-level NeRF training. ConceptGraph \cite{conceptgraphs} employs a greedy approach in the incremental process to associate the detection of the current frame with objects in the existing map. However, this straightforward method can lead to confusion between different objects that are in close proximity. To address this problem, we propose considering all frames together and devising a two-stage approach as shown in Fig. \ref{clustering}.

\textbf{Coarse Clustering Phase:} In this phase, we construct a mask graph $\mathcal{G}$ where each mask $m^{obj}_{t,i}$ is considered as a node. The weights between the masks are obtained from multiple similarities. The first is the geometric similarity $\textbf{S}_{geo}$, where each mask is projected into 3D space according to the corresponding depth $I^{d}_{t}[m^{obj}_{t,i}]$ to obtain a 3D point cloud. To facilitate fast matrix computation, we take the bounding box Intersection over Union (IoU) of the point cloud as $\textbf{S}_{pc}$. Next is the color similarity $\textbf{S}_{pho}$. The RGB three-channel color values of each mask are aggregated into a vector of $[32,3]$ and stitched together as a 1D color histogram. The final color similarity is obtained from the inner product of the color histograms. This is followed by the similarity of the two features, i.e., the cosine similarity $\textbf{S}_{clip}$ and $\textbf{S}_{cap}$ of the CLIP feature $f^{clip}_{t,i}$ and the caption feature $f^{cap}_{t,i}$. The final total weight matrix $\textbf{S}$ is obtained from the weighted sum of several similarity matrices:
\begin{equation}
\begin{aligned}
    \label{similarity}
    \textbf{S} = \omega_{geo}{\textbf{S}}_{geo}+\omega_{pho}{\textbf{S}}_{pho}+\omega_{clip}{\textbf{S}}_{clip}+\omega_{cap}{\textbf{S}}_{cap}
\end{aligned}
\end{equation}
where $\omega_{geo}+\omega_{pho}+\omega_{clip}+\omega_{cap}=1$, and $\textbf{S}$ is a large matrix of size $[N,N]$ ($N$ is the total number of all masks). 

For mask node pairs whose values in $\textbf{S}$ exceed the threshold ${{\theta }_{mask}}$, we add a similarity-weighted edge to the graph $\mathcal{G}$. For the obtained weighted undirected graph, the Louvain algorithm \cite{Louvain} is then applied to the resulting weighted undirected graph to cluster masks that belong to the same object. Since this method does not distinguish between the sources of the masks, it can effectively correlate masks across different frames and within the same frame, addressing issues of over-segmentation.

\textbf{Fine Clustering Phase:} Although the above coarse clustering phase effectively clusters the majority of masks belonging to the same object, some special cases remain. These exceptions are primarily due to objects being observed multiple times at the edges of images, making it difficult to integrate these parts into a cohesive whole. To address this issue, we perform the fine clustering phase. Using the results from the coarse clustering stage, we obtain a global point cloud and an averaged color histogram for each cluster. We then compute the matched points coverage rate and the similarity of color histograms between two clustered point clouds. The coverage rate indicates the proportion of matched points (with distances less than the threshold) among the lesser set.
If both metrics exceed thresholds ${{\theta }_{pc}}$ and ${{\theta }_{pho}}$, the two clusters are fused. For the final mask clustering result, if the number of elements in a cluster is less than a specified threshold $N/500$, the cluster is considered an outlier and discarded.

Each of the final clusters is considered as a mask collection $M({\mathcal{O}_{k}})$ of independent objects:
\begin{subequations}
\begin{align}
    \label{cluster_k}
     {\mathcal{C}_{k}} &= M({\mathcal{O}_{k}})\\ 
    \label{mask_k} 
    m_{t,j}^{{\mathcal{O}_{k}}} &\triangleq
    \{m_{t,j}^{obj}|m_{t,j}^{obj}\in {\mathcal{C}_{k}}\} 
\end{align}
\end{subequations}
where ${{\mathcal{C}}_{k}}$ is the $k$th cluster, ${{\mathcal{O}}_{k}}$ is the $k$th object instance and ${n}_{o}$ is the number of clusters (i.e., the number of objects).

\begin{figure}[!t]\centering
	\includegraphics[width=7cm]{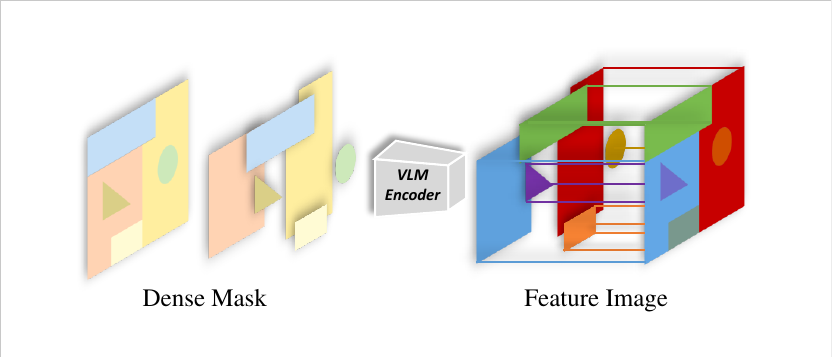}
	\caption{Part-level fine-grained feature extraction process: The mask $m^{part}_{t,j}$ extracted by SAM is dense and may be nested. The dense masks are visually encoded using VLMs, then averaged and superimposed to produce a feature image $I^{f}_{t}$ that matches the original image size.} 
	\label{partfeat} 
\end{figure}

\subsection{Part-level Fine-Grained Feature Extraction}

Both of the above modules operate at the instance level and do not perceive the interior details of the object. This can be insufficient in scenarios requiring fine-grained operations. To address this, the Part-level Fine-Grained Feature Extraction module is designed to generate dense feature images, which represent a refined, part-level understanding of the object. Fig. \ref{partfeat} visualizes this process.

Specifically, we apply SAM's automatic mask generation tool \cite{sam}, which has a powerful zero-sample 2D segmentation capability. Unlike the instance mask predictor CropFormer \cite{cropformer}, SAM segmentation generates masks with possible nesting between them, which can produce segmentation results with different granularities of objects. Taking the color images $I^{c}_{t}$ as input, SAM segments all the dense masks $\{m^{part}_{t,j} \mid j=1, 2, \ldots, n^{part}_t\}$. The images cropped along the edges of these masks are passed to CLIP to get the VLM feature vectors $f^{clip}_{t,j}$. We construct an empty image with the same size as the color image $I^{c}_{t}$ as an initialization of the feature image $I^{f}_{t}$. Next, we superimpose the features of these masks $m^{part}_{t,j}$ and perform normalization:
\begin{equation}
\begin{aligned}
    \label{part_feat}
    I_{t}^{f}=\frac{\sum\limits_{j}{\left( m_{t,j}^{part}\cdot f_{t,j}^{clip} \right)}}{\sum\limits_{j}{m_{t,j}^{part}}\ }
\end{aligned}
\end{equation}

In this manner, we generate dense feature images $I^{f}_{t}$ for each frame, akin to the color images $I^{c}_{t}$ and depth images $I^{d}_{t}$, which can then be utilized for subsequent NeRF training.

\begin{figure*}[!t]\centering
	\includegraphics[width=17.5cm]{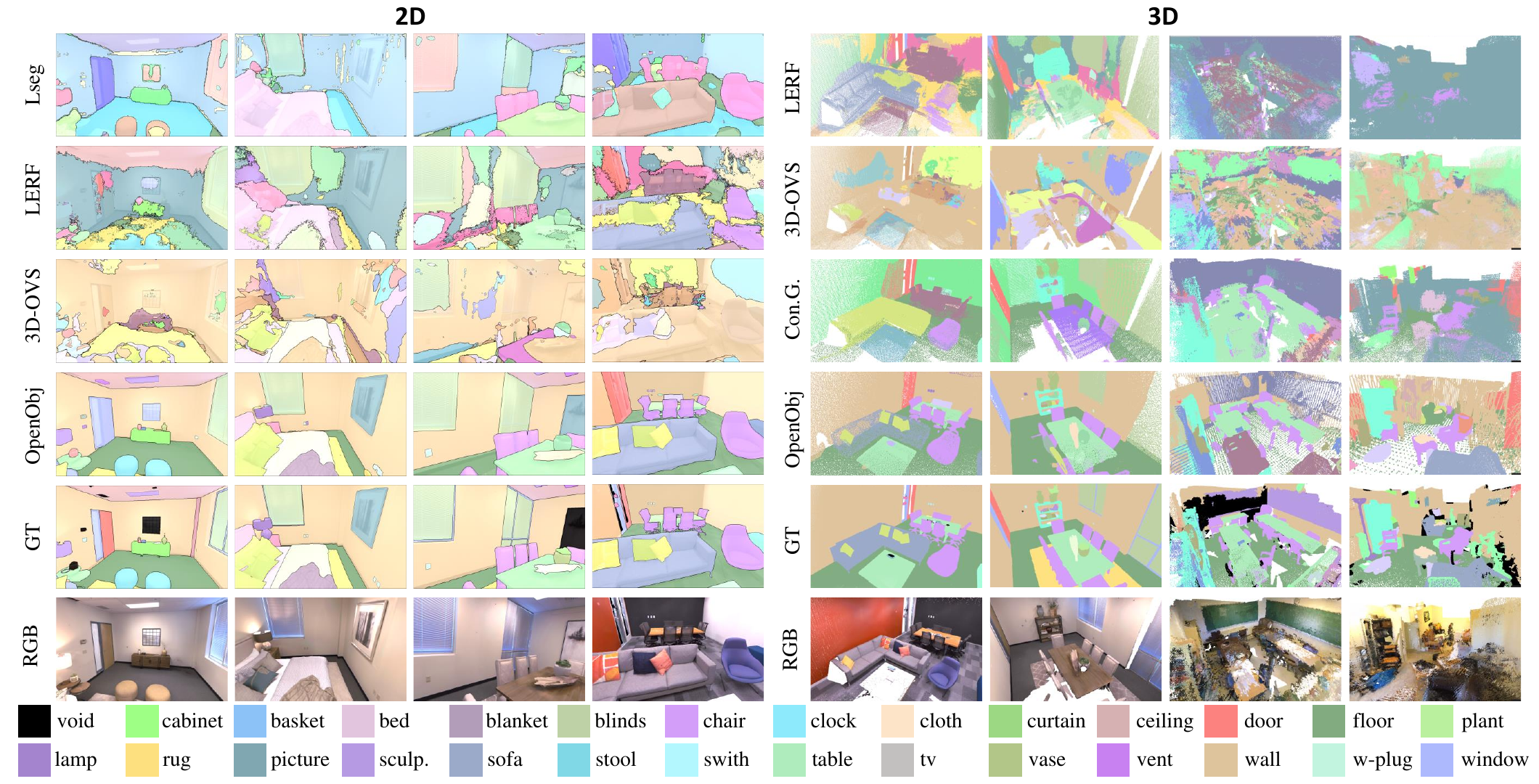}
	\caption{2D \& 3D zero-shot segmentation results. OpenObj's object-level NeRF and comprehensive understanding enable it to achieve clear boundaries and accurate semantics. } 
	\label{2d-3dss} 
\end{figure*}

\subsection{NeRF Rendering and Training}

In OpenObj, each object is modeled as a NeRF network with a uniform structure, enabling multi-model vectorized training similar to \cite{vmap}. We retain the 3D coordinate inputs $x, y, z$ while discarding the original direction inputs of NeRF. Additionally, we add output headers for feature vectors $f$ related to color $c$ and occupancy probability $o$, facilitating the learning of internal part understanding. Consequently, each object $\mathcal{O}_{k}$ can be represented as a NeRF network $\mathcal{F}_{\theta^k }^{k}$:
\begin{equation}
\begin{aligned}
    \label{mlp}
    \mathcal{F}_{\theta^k }^{k}:\{x,y,z\}\to \{c,o,f\}
\end{aligned}
\end{equation}
where $\theta^k$ denotes the network weights.

The entire rendering and training process is executed in the order of the image sequence. For the current frame images $\{I^{c}_{t}, I^{d}_{t}, I^{f}_{t}\}$, the object to which each instance mask $m^{\mathcal{O}_k}_{t,i}$ belongs has been determined as described in subsection \ref{MC}. Each object independently maintains $n_k$ keyframes, which serve as supervision for NeRFs. For each existing object $\mathcal{O}_{k}$, the specific supervision process is as follows:

\textbf{Sampling:} We first perform random pixel sampling to obtain the sampled pixels $[u,v]$. Using the camera intrinsic matrix $K$ and camera pose $P$, each pixel can be associated with a ray $\textbf{r}_{[u,v]}$ in the global coordinate system as $PK^{-1}[u, v]$.
We then perform $N_f$ and $N_s$ 3D point sampling on these rays, including $N_f$ uniform samples from the near boundary $t_n$ to the surface $t_s$ and $N_s$ normally distributed samples near the surface $t_s$. These sampled points are defined as $p_m$ and are ordered by the size of the depth value $d_m$. 

\textbf{Ray Rendering:} 
Feeding the NeRF with the aforementioned sampled points provides access to the color $c_m$, occupancy probability $c_m$, and features $f_m$ at the corresponding locations. These values are then rendered back into the image for 2D supervision. By parameterizing the occupancy probability to $[0,1]$, the termination probability of the ray $\textbf{r}_{[u,v]}$ at each point $p_m$ can be determined as ${{T}_{m}}={{o}_{m}}\prod\nolimits_{n<m}{(1-{{o}_{n}})}$. Based on this, we can render the occupancy, depth, color, and feature as:
\begin{equation}
\begin{aligned}
    \label{render}
   \hat{O}(\textbf{r}_{[u,v]})=\sum\limits_{m}{{{T}_{m}}}&, \quad \hat{D}(\textbf{r}_{[u,v]})=\sum\limits_{m}{{{T}_{m}}{{d}_{m}}} \\ 
  \hat{C}(\textbf{r}_{[u,v]})=\sum\limits_{m}{{{T}_{m}}{{c}_{m}}}&,\quad \hat{F}(\textbf{r}_{[u,v]})=\sum\limits_{m}{{{T}_{m}}{{f}_{m}}} \\ 
\end{aligned}
\end{equation}

\textbf{Loss Function:} Supervised training is conducted using the input images $\{I_{1:t}^{c}, I_{1:t}^{d}, I_{1:t}^{f}\}$. Pixel sampling is carried out within the 2D bounding box of the object ${\mathcal{O}_{k}}$ (denoted as $B({\mathcal{O}_{k}})$), while supervision of occupancy, depth, color, and feature is performed exclusively within the masks $M({\mathcal{O}_{k}})$:
\begin{subequations}
\begin{align}
     \mathcal{L}_{occ}^{k}&=\sum\limits_{[u,v] \in B({{\mathcal{O}}_{k}})}{\left| \hat{O}({{\textbf{r}}_{[u,v]}})-M({{\mathcal{O}}_{k}}) \right|}\\ 
     \mathcal{L}_{depth}^{k}&=M({{\mathcal{O}}_{k}})\cdot \sum\limits_{[u,v] \in B({{\mathcal{O}}_{k}})}{\left| \hat{D}({{\mathbf{r}}_{[u,v]}})-I^{d}[u,v] \right|}\\
     \mathcal{L}_{color}^{k}&=M({{\mathcal{O}}_{k}})\cdot \sum\limits_{[u,v]\in B({{\mathcal{O}}_{k}})}{\left| \hat{C}({{\mathbf{r}}_{[u,v]}})-{{I}^{c}}[u,v] \right|}\\
     \mathcal{L}_{feat}^{k}&=M({{\mathcal{O}}_{k}})\cdot \sum\limits_{[u,v]\in B({{\mathcal{O}}_{k}})}{\left| \hat{F}({{\mathbf{r}}_{[u,v]}})-{{I}^{f}}[u,v] \right|}
\end{align}
\end{subequations}
The overall loss function is obtained by summing the losses of all objects:
\begin{equation}
\begin{aligned}
    \label{all_loss}
    \mathcal{L}=\sum\limits_{k}{({{\lambda }_{1}}\mathcal{L}_{occ}^{k}+{{\lambda }_{2}}\mathcal{L}_{depth}^{k}+{{\lambda }_{3}}\mathcal{L}_{color}^{k}+{{\lambda }_{4}}\mathcal{L}_{feat}^{k})}
\end{aligned}
\end{equation}

\subsection{The Representation of OpenObj}

After completing the entire mapping process, the final map is represented as a set of object-level NeRFs along with an understanding of each object. The overall understanding of the object ${{\mathcal{O}}_{k}}$ is obtained by clustering the features of mask $m_{t,j}^{{\mathcal{O}_{k}}}$ belonging to that object and selecting the largest cluster as $f_{clip}^{{\mathcal{O}_{k}}}$ and $f_{cap}^{{\mathcal{O}_{k}}}$. This approach helps to mitigate the effects of outliers caused by poor observation viewpoints or model failures. 
Based on this, the overall VLM features $f_{clip}^{{\mathcal{O}_{k}}}$ and caption features $f_{cap}^{{\mathcal{O}_{k}}}$ facilitate open-vocabulary object retrieval, while the NeRF $\mathcal{F}_{\theta^k }^{k}$  enables fine-grained retrieval within the object. 

\section{Experimental Results}  \label{ER}
In this section, we aim to use experiments to validate OpenObj, through the following specific questions: 
\begin{enumerate}
  \item Without fine-tuning any model, can OpenObj achieve 2D and 3D segmentation of any scene with any class?
  \item Are OpenObj's open-vocabulary object-level and part-level retrieval results accurate and neat?
  \item What potential tasks can be facilitated by this multi-granularity representation?
\end{enumerate}

\begin{table}[!t]
\scriptsize
\renewcommand{\arraystretch}{1.1}
\centering
\caption{\textcolor{black}{2D Zero-shot Segmentation Results}}
\label{2dss_table}
\setlength{\tabcolsep}{0.5mm}{
{
\begin{tabular}{p{0.9cm}<{\centering}|p{0.8cm}<{\centering}p{0.8cm}<{\centering}p{0.88cm}<{\centering}p{0.95cm}<{\centering}|p{0.8cm}<{\centering}p{0.8cm}<{\centering}p{0.88cm}<{\centering}p{0.95cm}<{\centering}}
\toprule
\multirow{2}{*}{\textbf{Scene}} & \multicolumn{4}{c|}{\textbf{mIoU}} & \multicolumn{4}{c}{\textbf{mAcc}} \\ \cmidrule{2-9} 
 & \textbf{Lseg} & \textbf{LERF} & \textbf{3DOVS} & \textbf{OpenObj} & \textbf{Lseg} & \textbf{LERF} & \textbf{3DOVS} & \textbf{OpenObj} \\ \midrule \midrule
\textbf{room\_0} & 11.36 & 11.55 & 01.69 & \textbf{37.84} & 31.36 & 22.11 & 03.55 & \textbf{56.54} \\
\textbf{room\_1} & 09.00 & 14.90 & 00.73 & \textbf{33.39} & 30.72 & 34.47 & 01.72 & \textbf{54.81} \\
\textbf{room\_2} & 10.05 & 11.01 & 00.75 & \textbf{21.83} & \textbf{40.75} & 32.16 & 02.59 & 39.56 \\
\textbf{office\_0} & 07.82 & 06.45 & 01.12 & \textbf{21.36} & 27.40 & 13.29 & 07.90 & \textbf{40.29} \\
\textbf{office\_1} & 04.81 & 02.77 & 00.45 & \textbf{19.77} & 24.72 & 14.26 & 03.99 & \textbf{37.87} \\
\textbf{office\_2} & 08.47 & 08.20 & 00.55 & \textbf{13.30} & \textbf{29.49} & 20.65 & 01.87 & 25.75 \\
\textbf{office\_3} & 09.75 & 10.39 & 00.66 & \textbf{24.02} & 25.85 & 24.62 & 07.03 & \textbf{38.52} \\
\textbf{office\_4} & 07.00 & 09.01 & 01.01 & \textbf{25.61} & 34.11 & 29.20 & 08.34 & \textbf{47.10} \\ \midrule
\textbf{Average} & 08.53 & 09.28 & 00.87 & \textbf{24.64} & 30.55 & 23.85 & 04.62 & \textbf{42.56} \\ \bottomrule
\end{tabular}
}
}
\end{table}

\subsection{2D \& 3D Zero-shot Semantic Segmentation}

\textbf{Baseline:} 
For 2D semantic segmentation, we compare OpenObj with the language-driven image segmentation method LSeg \cite{LSeg}, as well as two state-of-the-art NeRF-based open-vocabulary mapping methods, LERF \cite{lerf} and 3D-OVS \cite{3dovs}, both of which construct point-wise feature fields. Except for LSeg, all other 2D segmentations based on NeRF methods are derived by rendering synthesized feature images. 
For 3D semantic segmentation, we add ConceptGraphs \cite{conceptgraphs} as a baseline to LERF and 3D-OVS, an open-vocabulary object-level point cloud map construction method. Considering that LERF and 3D-OVS do not introduce a depth prior, their results are obtained by projecting the 2D segmentation onto the ground-truth point cloud. In contrast, OpenObj's 3D point cloud is generated from the constructed surface of  $\mathcal{F}_{\theta^k }^{k}$.

\textbf{Datasets and Metrics:} 
We select two commonly used indoor datasets: eight scenes from Replica \cite{replica} and six scenes from ScanNet \cite{scannet}. Due to the lack of detailed 2D annotations in ScanNet, we opt to conduct 3D segmentation validation exclusively on the ScanNet dataset.
In the experiments, we use the semantic labels provided by the datasets as query text. For Replica, 101 classes are merged into 83 to combine very similar categories. These labels are encoded to obtain textual features. The similarity between the text features and the pixel or point features is calculated, with the highest similarity determining the semantic label.
For the evaluation metrics, we use mean IoU (mIoU) and mean accuracy (mAcc).

\begin{figure*}[!t]\centering
	\includegraphics[width=17.8cm]{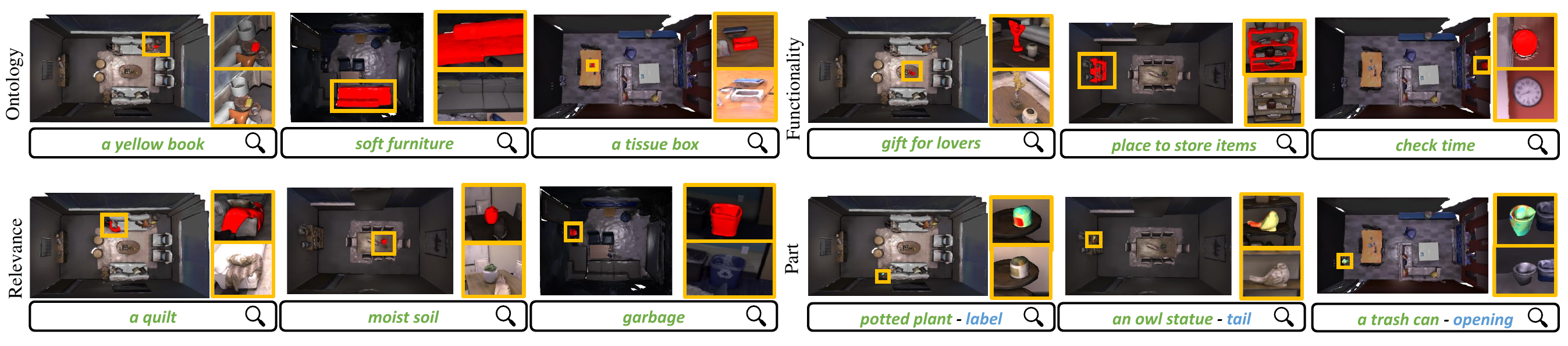}
	\caption{A selection of results from open-vocabulary retrieval. OpenObj correctly and clearly highlights the most relevant instance in each query.} 
	\label{retrieval} 
\end{figure*}

\begin{table}[!t]
\scriptsize
\centering
\caption{\textcolor{black}{3D Zero-shot Segmentation Results}}
\label{3dss_table}
\renewcommand\arraystretch{1.1}
\setlength{\tabcolsep}{0.5mm}
{
\begin{tabular}{p{1.0cm}<{\centering}|p{0.8cm}<{\centering}p{0.88cm}<{\centering}p{0.75cm}<{\centering}p{0.95cm}<{\centering}|p{0.8cm}<{\centering}p{0.85cm}<{\centering}p{0.75cm}<{\centering}p{0.95cm}<{\centering}}

\toprule
 & \multicolumn{4}{c|}{\textbf{mIoU}} & \multicolumn{4}{c}{\textbf{mAcc}} \\ \cmidrule{2-9} 

\multirow{-2}{*}{\textbf{Scene}} & \textbf{LERF} & \textbf{3DOVS} & \textbf{Con.G.} & \textbf{Ours} & \textbf{LERF} & \textbf{3DOVS} & \textbf{Con.G.} & \textbf{Ours} \\ \midrule \midrule
\textbf{room\_0} & 09.71 & 03.28 & 22.52 & \textbf{47.46} & 20.29 & 06.38 & 38.78 & \textbf{64.32} \\
\textbf{room\_1} & 22.09 & 01.10 & 18.74 & \textbf{39.60} & 36.62 & 03.04 & 36.59 & \textbf{56.53} \\
\textbf{room\_2} & 13.71 & 03.32 & 14.74 & \textbf{32.85} & 30.08 & 07.87 & 25.23 & \textbf{48.59} \\
\textbf{office\_0} & 07.19 & 01.63 & 19.35 & \textbf{26.09} & 13.90 & 07.90 & 29.30 & \textbf{42.14} \\
\textbf{office\_1} & 03.10 & 00.61 & 11.22 & \textbf{23.83} & 14.86 & 03.56 & 22.54 & \textbf{43.54} \\
\textbf{office\_2} & 09.04 & 00.87 & 15.79 & \textbf{16.98} & 20.48 & 05.15 & 33.78 & \textbf{30.89} \\
\textbf{office\_3} & 12.58 & 00.83 & 11.93 & \textbf{32.50} & 26.23 & 08.22 & 28.73 & \textbf{46.88} \\
\textbf{office\_4} & 12.33 & 03.74 & 17.52 & \textbf{29.60} & 33.09 & 14.24 & 37.26 & \textbf{48.45} \\ \midrule
\textbf{Average} & 11.22 & 01.92 & 16.48 & \textbf{31.11} & 24.44 & 07.05 & 31.53 & \textbf{47.67} \\ \midrule \midrule
\textbf{s.0011\_01} & 10.60 & 02.59 & 24.36 & \textbf{43.67} & 30.19 & 08.41 & 42.95 & \textbf{59.26} \\
\textbf{s.0030\_02} & 06.37 & 03.18 & 18.91 & \textbf{25.83} & 12.38 & 17.72 & 37.52 & \textbf{45.84} \\
\textbf{s.0220\_02} & 04.17 & 01.61 & 15.04 & \textbf{29.93} & 18.06 & 07.67 & 28.10 & \textbf{51.71} \\
\textbf{s.0592\_01} & 05.89 & 02.61 & 18.67 & \textbf{31.23} & 17.51 & 09.73 & 39.58 & \textbf{52.26} \\
\textbf{s.0673\_04} & 04.41 & 00.71 & 13.67 & \textbf{37.38} & 13.97 & 05.58 & 28.90 & \textbf{54.46} \\
\textbf{s.0696\_02} & 05.05 & 00.20 & 12.19 & \textbf{16.90} & 09.98 & 01.67 & 29.41 & \textbf{39.39} \\ \midrule
\textbf{Average} & 06.08 & 01.82 & 17.14 & \textbf{30.82} & 17.02 & 08.46 & 34.41 & \textbf{50.49} \\  \bottomrule
\end{tabular}
}
\end{table}

\textbf{Comparisons:} 
The qualitative and quantitative results of 2D zero-shot semantic segmentation are presented in Fig. \ref{2d-3dss} and Tab. \ref{2dss_table}, respectively. LSeg, as a fine-tuned model of CLIP, demonstrates the ability to capture pixel-aligned features, thereby showing sensitivity to object boundaries. However, this sensitivity comes at the cost of losing the capacity to recover complex concepts. On the other hand, both LERF and 3D-OVS, which are NeRF-based methods like OpenObj, employ a multi-scale or sliding-window technique to extract image features and utilize a single MLP to regress the feature field of the entire scene, resulting in a cluttered segmentation. In contrast, object-level NeRF-based OpenObj excels at distinguishing objects in the scene, naturally segmenting different instances. Moreover, OpenObj leverages the combined features of CLIP and caption, leading to a more comprehensive and robust understanding of objects.

The results of 3D zero-sample semantic segmentation are presented in Tab. \ref{3dss_table}. Similar to the 2D results, LERF and 3D-OVS encounter similar challenges. ConceptGraphs, which is an object-level open-vocabulary mapping method, fuses objects incrementally based on geometric and semantic similarities between different point cloud segments. However, this approach tends to under-segment the scene, as seen in Fig. \ref{2d-3dss} where the \textit{cushions} are merged with the \textit{sofa}. Additionally, relying solely on CLIP-based features can result in inaccurate object classification.
In contrast, OpenObj adopts a two-stage mask clustering approach, which leads to a more optimal solution for global object segmentation,  and combines object understanding to achieve accurate 3D semantic segmentation.

\subsection{Multi-granularity Open-vocabulary Retrieval}

\textbf{Baseline:} For the retrieval experiments, only ConceptGraphs \cite{conceptgraphs} with object-level concepts is kept as a baseline.

\textbf{Datasets and Metrics:} The experiments are conducted on four scenes in Replica \cite{replica}, each featuring a diverse array of objects. We categorized the retrieved text into four types, with the first three being object-level retrievals:
\begin{itemize}
\item Ontology: Directly describe the characteristics of the object itself, For example, \textit{a brown sofa}.
\item Relevance: Discuss elements related to the object. For example, \textit{airflow} (that is  \textit{vent}).
\item Functionality: Emphasize the function of the object. For example, \textit{garbage collection} (that is \textit{trash can}).
\item Part: Focus on describing both the object and its internal parts. For example, \textit{a wooden door - handle}.
\end{itemize}
For each scene, we selected five samples for each retrieval type. To evaluate the performance, we measure the recall at the top-1, top-2, and top-3 levels.

\begin{table}[]
\centering
\scriptsize
\caption{Retrieval Results (top-1,2,3 recall)}
\label{retrieval_table}
\renewcommand\arraystretch{1.1}
\setlength{\tabcolsep}{1.5mm}{
\begin{tabular}{cccccc}
\toprule
\textbf{Retrieval-Type} & \textbf{Methods} & \textbf{R@1} & \textbf{R@2} & \textbf{R@3} & \textbf{\#Retrieval} \\ \midrule
 & Con.G. & 0.65 & 0.70 & 0.70 &  \\
\multirow{-2}{*}{\textbf{Ontology}} & OpenObj & \textbf{0.90} & \textbf{0.95} & \textbf{1.00} & \multirow{-2}{*}{20} \\ \midrule
 & Con.G.  & 0.50 & 0.70 & 0.80 &  \\
\multirow{-2}{*}{\textbf{Relevance}} & OpenObj & \textbf{0.75} & \textbf{0.90} & \textbf{1.00} & \multirow{-2}{*}{20} \\ \midrule
 & Con.G. & 0.50 & 0.80 & 0.85 &  \\
\multirow{-2}{*}{\textbf{Functionality}} & OpenObj & \textbf{0.90} & \textbf{0.95} & \textbf{0.95} & \multirow{-2}{*}{20} \\ \midrule
 & Con.G. & - & - & - &  \\
\multirow{-2}{*}{\textbf{Part}} & OpenObj & \textbf{0.80} & \textbf{0.80} & \textbf{0.80} & \multirow{-2}{*}{20} \\ \bottomrule
\end{tabular}
}
\end{table}

\textbf{Comparisons:}  The quantitative results of the retrieval experiments are presented in Tab. \ref{retrieval_table}, with some of OpenObj's retrieval results illustrated in Fig. \ref{retrieval}. OpenObj consistently outperforms ConceptGraphs across all types of retrieval tasks. This superior performance can be attributed to the introduction of the caption feature, which provides OpenObj with more direct references and enhances its semantic inference about objects. 
As for part retrieval, ConceptGraphs, limited to object-level understanding, are incapable of accomplishing this task. Conversely, OpenObj demonstrates a marked advantage in handling patterns, components, and other parts. 

\begin{figure}[!t]\centering
	\includegraphics[width=7.8cm]{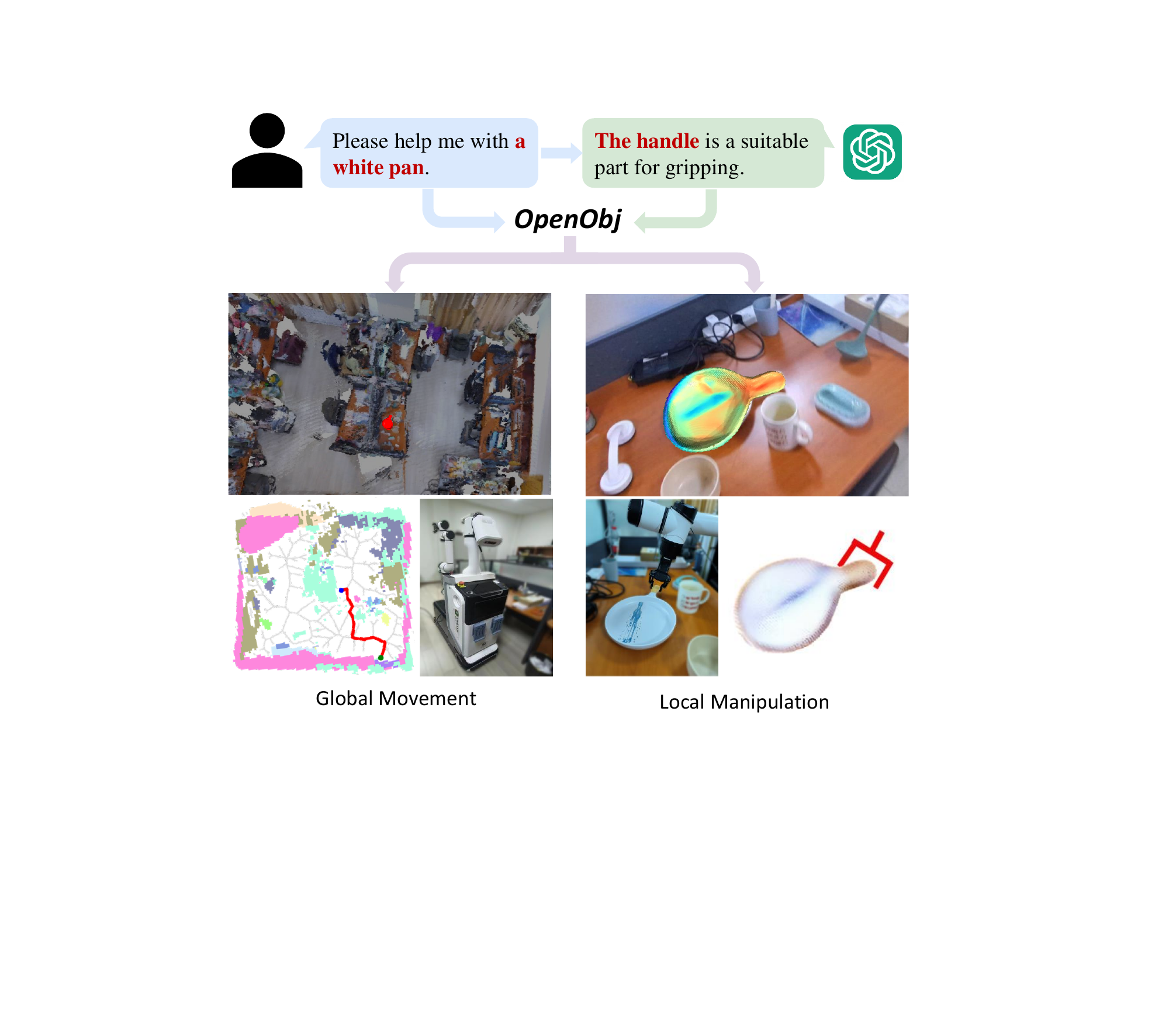}
	\caption{OpenObj's multi-granularity scene understanding supports multi-granularity downstream tasks, including object-oriented global movement and part-oriented local manipulation.} 
	\label{demo} 
\end{figure}

\subsection{Global Movement and Local Manipulation}

\textbf{The task:}  
OpenObj's distinctive representation enables it to support a range of downstream tasks at different levels of granularity. These tasks include object-oriented global movement and part-oriented local manipulation. To validate this capability, we conducted an experiment involving a mobile robot equipped with a robotic arm in an office.  The user can issue a find object command to the robot, and the robot's objective is to navigate to the specified location and execute the appropriate grasping action.

\textbf{Implementation details:} 
For global navigation, the process commences by generating a 2D obstacle grid map based on predefined height. Subsequently, a Voronoi graph is constructed on this map to establish a feasible path for the robot. The final navigation trajectory is determined by identifying the most relevant objects in the map.
Regarding local manipulation, the initial step involves utilizing GraspNet \cite{Graspnet} to generate potential grasping poses for the object. Subsequently, the ChatGPT interface \cite{GPT4} is employed to identify a suitable grasping part. This textual description of the grasping part is then utilized for fine retrieval within the object, combining its similarity score with GraspNet's grasping score to determine the optimal grasping pose.

\textbf{The results:} 
Fig. \ref{demo} showcases the outcome of one such process. 
In response to the user's command, \textit{`Please help me with \uline{a white pan}'}, OpenObj retrieves the target object and devises a secure navigational path for the robot. With guidance from ChatGPT, which identifies the most appropriate part (\textit{` \uline{the handle}'}), the optimal grasping pose is determined by considering the current candidate poses. Finally, the robot completes the grasping action successfully .

\section{Conclusion} \label{C}

This paper introduces OpenObj, an innovative approach to build open-vocabulary object-level Neural Radiance Fields with fine-grained understanding. 
First, OpenObj segments and interprets object instances from the input color image. Subsequently, a two-stage mask clustering approach is employed to achieve cross-frame object associations. SAM over-segmentation properties then aid in constructing feature images. Finally, vectorized training of all objects' NeRF models is accomplished through multi-loss supervision. Multiple experiments validate the advantages of OpenObj in zero-shot semantic segmentation and open-vocabulary retrieval. Furthermore, mobile manipulation experiments demonstrate the applicability of OpenObj for potential downstream tasks.

\bibliographystyle{Bibliography/IEEEtran}
\bibliography{Bibliography/RAL}

\end{document}